\documentclass[acmlarge]{acmart}

\usepackage{booktabs} 
\usepackage{tabu}
\usepackage{makecell} 
\usepackage[ruled]{algorithm2e} 

\SetAlFnt{\small}
\SetAlCapFnt{\small}
\SetAlCapNameFnt{\small}
\SetAlCapHSkip{0pt}
\IncMargin{-\parindent}

\newenvironment{conditions}[1][where:]
  {#1 \begin{tabular}[t]{>{$}l<{$} @{${}={}$} l}}
  {\end{tabular}\\[\belowdisplayskip]}

\setcopyright{usgovmixed}

\begin{document}
\title{A Causality-Aware Pattern Mining Scheme for Group Activity Recognition in a Pervasive Sensor Space}

\author{Hyunju Kim}
\affiliation{%
  \institution{Korea Advanced Institute of Science \& Technology}
  \city{Daejeon}
  \country{South Korea}}
\email{iplay93@kaist.ac.kr}

\author{Heesuk Son}
\affiliation{%
  \institution{Meta}
  \city{Washington}
  \country{United States}}
\email{heesuk.chad.son@gmail.com}

\author{Dongman Lee}
\affiliation{%
  \institution{Korea Advanced Institute of Science \& Technology}
  \city{Daejeon}
  \country{South Korea}}
\email{dlee@kaist.ac.kr}

\begin{abstract}
Human activity recognition (HAR) is a key challenge in pervasive computing and its solutions have been presented based on various disciplines. Specifically, for HAR in a smart space without privacy and accessibility issues, data streams generated by deployed pervasive sensors are leveraged. 
In this paper, we focus on a group activity by which a group of users perform a collaborative task without user identification and propose an efficient group activity recognition scheme that extracts causality patterns from pervasive sensor event sequences generated by a group of users to support as good recognition accuracy as the state-of-the-art graphical model with missing or false data tolerance. To filter out irrelevant noise events from a given data stream, a set of rules is leveraged to highlight causally related events. Then, a pattern-tree algorithm extracts frequent causal patterns by means of a growing tree structure. Based on the extracted patterns, a weighted sum-based pattern matching algorithm computes the likelihoods of stored group activities to the given test event sequence by means of matched event pattern counts for group activity recognition. We evaluate the proposed scheme using the data collected from our testbed and 
CASAS datasets where users perform their tasks on a daily basis and validate its effectiveness in a real environment. Experiment results show that the proposed scheme performs higher recognition accuracy and is tolerant to missing or false data with a smaller amount of runtime overhead than the existing schemes. 

\end{abstract}

%
%

\ccsdesc[500]{Human-centered computing~Ubiquitous and mobile computing design and evaluation methods}
\ccsdesc[300]{Information systems~Information systems applications}

%
%

\keywords{group activity recognition, causality-aware, pattern mining, pervasive sensors, smart space}


\maketitle


\section{Introduction}

Human Activity Recognition (HAR) is a fundamental technology to provide appropriate services to users in a smart space \cite{Benmansour2016,Tapia2004}. 
It is achieved by exploiting various sensors available in a given space. One of approaches is to leverage video surveillance or wearable sensors. It is used in many HAR schemes but has to 
face several issues such privacy, accessibility, unstable data collection or battery \cite{Pantelopoulos2010, Foresti2012, Al2012}. 
In order to recognize activities while relieving such issues in a smart spaces,
data streams generated by deployed IoT devices (smart objects) such as pervasive sensors (e.g. sound, temperature, etc.) and actuators (beam projector, air conditioner, etc) have been leveraged. Using multivariate data streams, early studies have focused on finding simple activities which consist of each user's sequential sensor events. To analyze simple activities, various graphical models such as Hidden Markov Model (HMM) have been presented \cite{chen2014,Wang2011,Chiang2010,zhang2006}. 
However, 
people often perform not only simple activities but also
complex activities that encompass 
various events at the same time within single user or multi-user activities. 
Therefore, current studies have attempted to recognize complex activities by representing rich temporal relationships (e.g. overlap, concurrent) between events \cite{riboni2016,liu2016,Benmansour2016,lotfi2016}.

Considering the necessity of handling concurrent sensor events in a complex activity recognition problem, improved models with Allen's temporal relations \cite{allen1994} are devised \cite{zhang2013,liu2016,lotfi2016}. 
Interval Temporal Bayesian network (ITBN) \cite{zhang2013} represents complex temporal relationships among sensor events onto the edges of a Bayesian network. Liu et al. \cite{liu2016} advance the ITBN by leveraging the Chinese Restaurant Process (CRP) \cite{aggarwal2012} for ITBN constructions to describe repetitive events. Despite their expression power, the graphical models calculate the transition probability of all variations of event sequences for each complex activity (called a complex activity model) and compute how closely a given event sequence matches with each complex activity model. This gives a rise to low recognition accuracy if target event sequences include those which do not exactly match with the corresponding complex models. The limitation significantly affects the recognition of a complex activity induced by a group of users together without user identification (hereafter, we call it a \textit {group activity}). Since a group activity (e.g. meeting, discussion, etc.) entails concurrent sub-tasks by multiple anonymous users, it is hard to clearly detect any relationships among sensor events (e.g. sound level, projector on/off, etc.), users, and sub-tasks (i.e. data association problem \cite{hsu2008}). Actually, Liu et al. \cite{liu2016} derive a great improvement on recognition accuracy with a dataset \cite{opportunity} which is composed of a single user-driven complex activities. However, the accuracy decreases when applied to a group activity dataset \cite{OSUPEL}.
To relieve the fixed structure of graphical models, the association rule mining-based approaches \cite{lotfi2016,Gu2009,Liu2015} extract representative event associations from training event sequences and infer complex activities of target event sequences. They allow a given event sequence to be correctly recognized as long as it includes the representative event associations trained for the corresponding complex activity. Bourbia et al. \cite{lotfi2016} solve the concurrent event representation issue by adopting Allen's temporal relation model and prove their scheme is valid for group activity recognition. However, they ignore the temporal orders, causality, of disjunct sensor events considered in the graphical models for the ease of computation, which discards important patterns in the mining phase and degrades the group activity recognition accuracy eventually. 

In this paper, we propose a causality-aware group activity recognition scheme that exploits the applicability of pattern mining and achieves high recognition accuracy by preserving potential temporal event orders (i.e. causality) in complex activities. For this, we devise two methods for extracting event causality patterns. First, we define a set of event causality in terms of ontology that exists in group activities (GAs). If events with temporal precedence are detected, ontology-based rules derive causality between events. Second, we leverage sequential pattern mining method \cite{Salvemini2011} for extracting additional event causality that are not detected by the first method but frequently generated in given GA sequences. The extracted patterns are represented by associated first order logic and their causalities. In addition, we devise a method for handling infrequent but important GA-specific events from smart objects such as lights and projectors which, otherwise, are ignored in mining phase due to their relatively low event frequency. To evaluate the proposed scheme, we perform a leave-one-out cross validation and compare the proposed scheme with three existing schemes (Association rule mining \cite{lotfi2016}, CRP-based ITBN \cite{liu2016} and HMM \cite{Eddy1996})  in terms of recognition accuracy, training performance and noise sensitivity. For a dataset, we leverage the group activity IoT stream data which has been collected for 
14 months from our testbed. Experiment results show that the proposed scheme improves the recognition accuracy by 
21\%, 19\%, and 30\% compared to association rule mining scheme, CRP-based ITBN scheme and HMM scheme, respectively, considering 
episodes where data is missing or false values are fed due to errors or malfunctions of some pervasive sensors or actuators.
In addition, we prove the scalability (i.e. works well in other environments) of our scheme using the CASAS group activity dataset \cite{CASAS} and the result shows that our scheme improves at least 12\% recognition accuracy rate than the existing schemes. 


\pagebreak
\section{Related Work}
%
In order to recognize human activities in a smart space, wearable sensors or video - based approaches are proposed. However, those disciplines have privacy or accessibility issues to use in daily living, data streams generated by deployed pervasive sensors (e.g. sound, seat occupancy, etc.) without user identification have been leveraged. While early approaches \cite{chen2014,Wang2011,Chiang2010,zhang2006} studied the recognition of simple activity which consists of sequential events, recent studies \cite{riboni2016,liu2016,Benmansour2016,lotfi2016} have focused on recognizing complex activity composed of events with various temporal relations.

Markov property-based models \cite{riboni2016,chen2014,Wang2011,Chiang2010} are the most widely leveraged approaches for the complex activity recognition domain. However, they have limitations of describing temporal relations between events. 
Riboni et al \cite{riboni2016} propose the unsupervised and knowledge based model using ontology. They derive semantic correlations among activities and events to represent the characteristics of complex activities. This allows their scheme to be exploited to various environments. However, the scheme cannot be directly applied to a pervasive sensor space because it is difficult to manually pre-define all sensors and the probabilistic relations of group activities. It has a limitation to represent temporal dependencies between events such as overlapping or concurrent relations which frequently exist in a pervasive sensor space, resulting in accuracy degradation. The other Markov property-based approaches mentioned above also have the same limitation.

Zhang et al. \cite{zhang2013} present Interval Temporal Bayesian Network (ITBN) \cite{zhang2013} which is an advanced graphical model with a richer representation, Allen's temporal relations model \cite{allen1994}, to describe temporal dependencies between states (e.g. complex activity events). As with other graphical models, ITBN represents events as vertices and puts their temporal relations on in-between edges. However, ITBN suffers from an expensive computational cost and a fixed network size, which makes it impossible to represent repetitive occurrences of events. Liu et al. \cite{liu2016} relieve these limitations by clustering similar complex activity patterns as multiple ITBNs with different network sizes based on the Chinese Restaurant Process (CRP). Despite its expression power and high accuracy, the inherently exact matching process of the graphical model makes itself vulnerable to changes of user behaviors. Especially, various patterns of user behaviors for the same GA occur in a pervasive sensor space and it makes hard to secure enough dataset for the same complex activity. In this aspect, the accuracy of the trained ITBN model can be degraded even if behaviors of users slightly differ from the trained ones (i.e. missing or false data tolerance). It is more critical to recognize complex activity driven by group (group activity) since GA entails concurrent sub-tasks (e.g., meeting, discussion, etc.) by multiple anonymous users with no user identification (i.e. data association problem \cite{hsu2008}). The results of Liu et al.  \cite{liu2016} show recognition rates of a group activity dataset are much lower than recognition rates of a single user driven complex activity dataset. 

To construct a noise-robustness model in a pervasive sensor space, association rule based approaches \cite{lotfi2016,Gu2009,Liu2015} are leveraged to model complex activity recognition module. Bourbia et al. \cite{lotfi2016} represent temporal relationships of sensor events as the first order logic and extract their patterns as emerging patterns. They extract patterns composed of sensor events with frequent co-appearances based on association rule mining \cite{agrawal1993}. In the proposed scheme, temporal relations between sensor events are represented by Allen's temporal relation model and frequent relations are extracted as GA patterns. The extracted patterns are encoded together as a Markov Logic Network (MLN) instance and MCMC-based inference is performed to recognize a given test event sequence. Their results show valid accuracy results in GAs, however, they overlook an important aspect of the mined patterns, that is, causality embedded in each GA and thus fail to achieve as high recognition accuracy as the graphical model.

\pagebreak
\section{Design Considerations}
In this section, we discuss key considerations to overcome the limitations of the existing schemes, that is, support high accuracy and low noise sensitivity for group activity recognition (GAR).

\subsection{Filtering out non-causal noise events}

Extracting GA patterns using machine learning techniques is inherently based on frequency. However, this characteristic seriously degrades the learning accuracy in a smart space with various pervasive sensors because those sensors may produce a stream of thousands of events for a single GA while only a few of them embody meaningful patterns. It is highly probable that noise events which have no causal relation form frequent patterns and these patterns are derived as important ones for a target GA. Furthermore, as the amount of stream data grows, a learning phase may suffer from memory shortage. To prevent these problems and secure the GA recognition accuracy, it is necessary to filter out semantically irrelevant event pairs and detect only those that may affect each other's appearance (i.e. causality) from the pipelined event streams .

\begin{figure}[!b]

  \centerline{\includegraphics[scale=0.5]{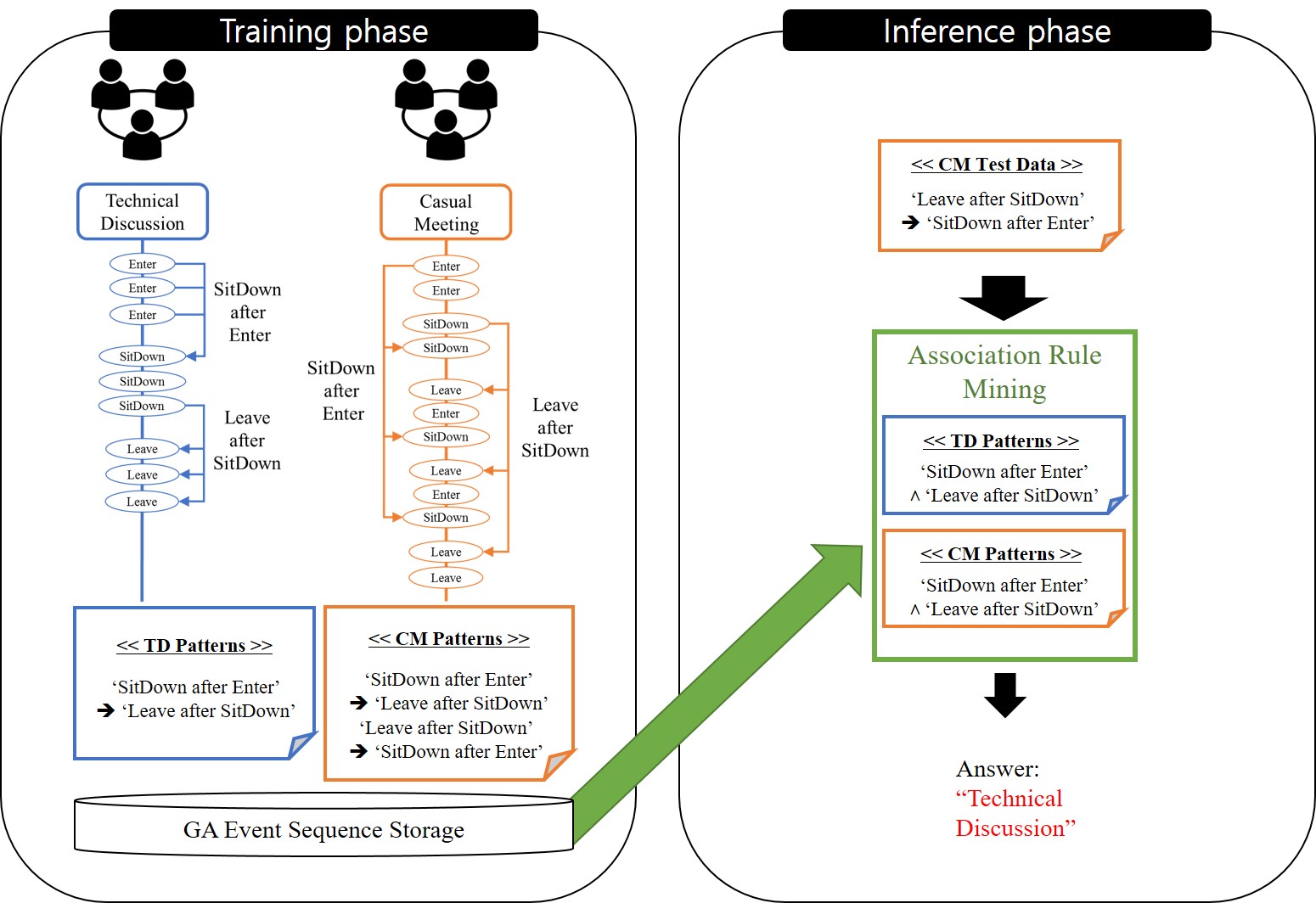}}

  \caption{ A limitation of association rule mining and a need for considering temporal dependency for better GAR(Group Activity Recognition)}

  \label{fig1}

\end{figure}

\subsection{Preserving causality within patterns }

In a pervasive sensor space, similar types of event pairs can be generated in different GAs. The key attribute to differentiate those GAs is a sequential variation of event pair combinations. For example, when project team members have a monthly meeting to have a technical discussion, they usually enter a meeting room, sit, discuss with little disruptions, and leave the room together. On the other hand, when they have an informal gathering for smalltalk or coffee break, they may freely leave and enter a room for many reasons (e.g. toilet, phone call, material print-out, etc). In both cases, common event pairs such as \textit {Leave after SitDown} and \textit {SitDown after Enter} are generated repeatedly. However, their generation patterns are different as illustrated in Figure \ref{fig1}. Existing association rule mining schemes \cite{lotfi2016} cannot classify this difference, resulting in high false negatives as well as false positives in the GA recognition phase. To overcome this limitation and improve the recognition accuracy, we need to devise a richer representation method that is capable of separating different temporal orders between event pairs.

\subsection{Saving infrequent but important GA-specific events}

Two types of smart objects are usually deployed in a pervasive sensor space: pervasive sensors and actuators. Pervasive sensors may publish thousands of events (i.e. sensed environmental raw data or their discretized values) in an hour, while actuators (e.g. a smart light bulb, a projector, etc.) publish a few state changes, mostly done by user manipulations. While actuators have important roles in performing GAs, actuator event patterns are hardly extracted by machine learning-based GA recognitions due to such unbalanced event generation. For example, projector manipulation events, in a technical discussion using a projector, are hardly drawn as patterns than sound level events because the former happens much less than the latter. To preserve such infrequent but important GA-specific events, there needs a way to identify them and treat them as equal as frequent ones.

\section{Proposed Scheme}

\begin{figure*}[htbp]
  \includegraphics[scale=0.22]{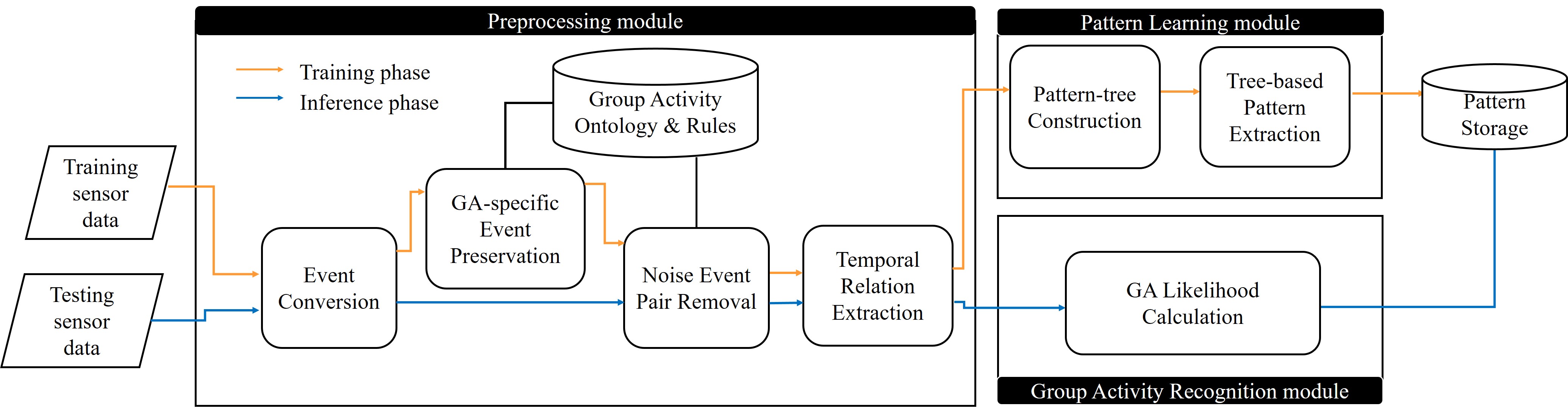}
  \caption{Overall architecture of proposed scheme}
  \label{fig2}
\end{figure*}

\subsection{Overview }
In this section, we describe how the proposed scheme fulfills the aforementioned key requirements and extracts causality-aware patterns from group activity (GA) event sequences. Figure \ref{fig2} illustrates its key components and the operation flows for training and inferencing pervasive sensor data. The preprocessing module translates sensor date into a sequence of events representing a group activity, finds GA-specific events and event causality by the group activity ontology from it, and produces a set of temporal relations defined in Allen's model \cite{allen1994}. Then, the pattern learning module constructs a pattern-tree which computes sequential orders between the temporal relations and detects frequent causality patterns of a target group activity. The derived patterns are stored in the pattern storage. For a test event sequence to inference, the preprocessing module extracts temporal relations from it as the training phase and the group activity recognition (GAR) module infers the most proper group activity. For this, the GAR module compares given temporal relations with the learned GA patterns in the pattern storage and compute each GA's likelihood in terms of a weight sum. A GA with a largest weight sum is selected as a recognition output.

\subsection{Preprocessing Module}
\subsubsection{Event Conversion} 
At first, when a raw sensor data sequence comes in as an input, it is transformed into an event sequence. Raw data is composed of a value of a sensor and its published timestamp, while event data consists of its start time, end time and label. Two types of objects usually exist in a pervasive sensor space: actuators (e.g. Seat or Projector) and pervasive sensors (e.g. Lightness or Sound). Actuator data is generated when a user's action affects an actuator's state change. Thus, its state value is directly converted to an event label. The times when a person actuates and ends an actuator correspond to the start time and end time of an actuator event, respectively. However, an pervasive sensor requires additional data processing to convert its states to an event level since its states can change instantaneously and can contain many noises. We apply a simple signal averaging method to the raw data of pervasive sensors. We convert each sampled raw data into a human-understandable level (e.g. raw sound data \lq 56\rq to \lq Sound\_Level1 \rq). If the same level of data occur consecutively, they are considered the same event and the time at which a different level of data appears is the end time of the event. After completing this step, we can get all event tuples of data consisting of start time, end time, and event label.
\begin{figure}[H]
\begin{minipage}[t]{0.47\linewidth}
    \includegraphics[width=\linewidth]{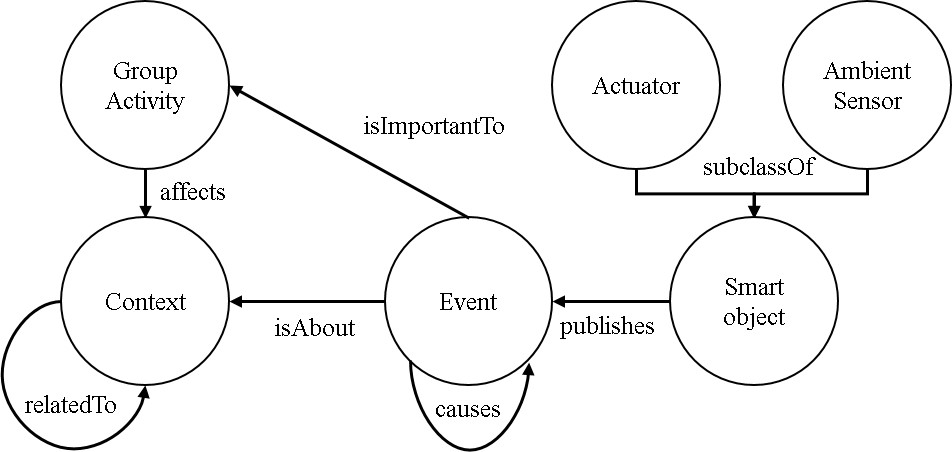}
    \caption{Group activity ontology}
    \label{fig3}
\end{minipage}%
    \hfill%
\begin{minipage}[t]{0.5\linewidth}
    \includegraphics[width=\linewidth]{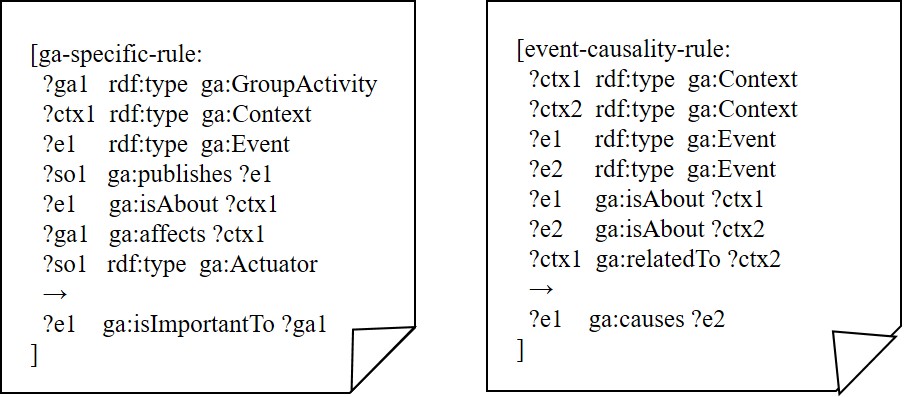}
    \caption{Event filtering rules in GA knowledge base}
    \label{fig4}
\end{minipage} 
\end{figure}
\subsubsection{Group Activity Knowledge base Definition \& Construction} 
To detect GA-specific events and event pairs having causality from the event streams, we design a group activity knowledge base containing an ontology and rules. 

Group activity ontology (Figure \ref{fig3}) defines GA-related key concepts and their relationships. Key terms are defined as follows:
\newline
\begin{itemize}
  \item Group Activity (e.g. Technical Discussion) : A task that multiple users perform together in a smart space.  
  \item Context (e.g. ProjectorStatus) : A human-understandable term of a smart space component.
  \item Event (e.g. ProjectorOn) : A specific state \lq value \rq of a smart object.
  \item Smart Object (e.g. Projector) : An IoT device that comprises a smart space.
  \newline
\end{itemize}

Then, we define a predicate, \textit{affects}, to declare \textit{Contexts} which are affected by a specific \textit{Group Activity} in the knowledge base. We define another predicate, \textit{relatedTo}, to declare causality between \textit{Contexts}. A set of ontology triples below shows an example GA knowledge base we built for our smart space 
(`ga' is the name space prefix of the GA ontology) :

\begin{table}[h]
\begin{center}
\begin{tabular}{ll}
[ga:TechnicalDiscussion-ga:affects-ga:SoundLevel]\\[1.5 mm]
[ga:TechnicalDiscussion-ga:affects-ga:ProjectorStatus]\\[1.5 mm]
[ga:Projector - ga:publishes - ga:ProjectorOn]\\[1.5 mm]
[ga:ProjectorOn - ga:isAbout- ga:ProjectorStatus]\\[1.5 mm]
[ga:Projector - rdf:type - ga:Actuator]\\[1.5 mm]
[ga:Sound - ga:publishes - ga:SoundLevel1]\\[1.5 mm]
[ga:Sound - ga:publishes - ga:SoundLevel2]\\[1.5 mm]
[ga:SoundLevel1 - ga:isAbout- ga:SoundLevel]\\[1.5 mm]
[ga:SoundLevel2 - ga:isAbout- ga:SoundLevel]\\[1.5 mm]
[ga:Sound  - rdf:type - ga:PervasiveSensor]\\[1.5 mm]
[ga:ProjectorStatus - ga:relatedTo - ga:SoundLevel]\\[1.5 mm]

\end{tabular}
\end{center}
\end{table}%

When constructing the knowledge base above, statements with \textit{affects} and \textit{relatedTo} predicates are supposed to be written by those who have an overall knowledge on a target smart space (e.g. space administrator). Statements with other predicates such as \textit{publishes} and \textit{isAbout} are supposed to be written by smart object developers. However, we cannot assume that an administrator always exists in a smart space. Then, it would be difficult to figure out which \textit{Contexts} correspond to \textit{affects} and \textit{relatedTo} relations without help of an administrator. To overcome this problem, it is required to automatically find important \textit{Contexts} and causalities to each group activity based on the training data. For that purpose, we exploit TF-IDF \cite{book-salton} which is a widely used weighting factor to reflect how important a word is to a document in information retrieval domain. 
To discover \textit{affects} relations, the importance of \textit{Contexts} within each \textit{Group Activity} is calculated as equation \ref{eqone}. Then, if the result is non-zero, the corresponding \textit{Context} is considered to be affected by that \textit{Group Activity}.

\begin{equation}
I_{c_{x,z}} = f_{c_{x,z}} \times \log_{10}{\frac{N}{gaf_{x}}}
\label{eqone}
\end{equation}

\begin{conditions}
 f_{c_{x,z}}  &  ratio of \textit{Context} $x$ in \textit{Group Activity} $z$\\
 gaf_{x}    &  number of group activities containing \textit{Context} $x$ \\
 N &  total number of group activities
\end{conditions}

Finding \textit{relatedTo} relations among \textit{Contexts} is similar. First, two \textit{Contexts} with a sequential relation are converted into an ordered set and the number of each ordered set is counted. Equation \ref{eqtwo} calculates the importance of each ordered set in each \textit{Group Activity}.

\begin{equation}
I_{os_{(x,y),z}} = f_{os_{(x,y),z}} \times \log_{10}{\frac{N}{gaf_{(x,y)}}}
\label{eqtwo}
\end{equation}

\begin{conditions}
(x,y) & an ordered set implying that \textit{Context} $x$ occurs before \textit{Context} $y$   \\
f_{os_{(x,y),z}}  &  ratio of an ordered set $(x,y)$ in \textit{Group Activity} $z$ \\
gaf_{(x,y)}    &  number of group activities containing an ordered set $(x,y)$ \\
 N &  total number of group activities
\end{conditions}

\begin{equation}
\min_{z \in Z}  \{ I_{os_{(x,y),z}}\}~~\text{~~~~~~~~~~~($Z$ is a set of all group activities)}
\label{eqthree}
\end{equation}

If the result of equation \ref{eqthree} is zero, it implies that an ordered set of \textit{Context} $x$ and $y$ is not related to any GAs at all. We filter out such an ordered set regarded as non-causal noise context relations while \textit{Context} pairs in the ordered sets not filtered are stored in the knowledge base in the form of statements with \textit{relatedTo} predicate. 

Based on the terms defined in the ontology, we create the \textit{ga-specific-rule} to examine if an incoming event sequence contains GA-specific events and the \textit{event-causality-rule}  to detect event pairs having causality. Both logic are written in n3 format \cite{n3format} and stored in GA knowledge base (Figure \ref{fig4}).

\subsubsection{GA-specific Event Preservation}\label{ga-specific} 
At first, we derive GA-specific events from an event stream based on the stored triples and \textit{ga-specific-rule} of the GA knowledge base. An event having  \textit{isImportant}, relationship with a particular GA is drawn as a GA-specific event. The following example shows how the rule determines if \textit{ProjectorOn} is a GA-specific event in \textit{TechnicalDiscussion} :
\begin{table}[H]
\begin{center}
\begin{tabular}{ll}
$[$ ga-specific-rule:\\[1.5 mm]
     TechnicalDiscussion - rd:type - ga:GroupActivity \\ [1.5 mm]
     ProjectorStatus - rd:type - ga:Context\\ [1.5 mm]
     ProjectorOn - rd:type - ga:Event\\[1.5 mm]
     Projector - ga:publishes - ProjectorOn\\[1.5 mm]
     ProjectorOn - ga:isAbout - ProjectorStatus\\[1.5 mm]
     TechnicalDiscussion - ga:affects - ProjectorStatus\\[1.5 mm]
     Projector - rdf:isTypeof - ga:Actuator\\[1.5 mm]
   $\longrightarrow$ ProjectorOn - ga:isImportantTo - TechnicalDiscussion\\[1.5 mm]
$]$

\end{tabular}
\end{center}
\end{table}%

\textit{ProjectorOn}  can be a GA-specific event of \textit {TechnicalDiscussion}. If a GA-specific event is detected, the preprocessing module emphasizes its importance by splitting the event into N times events. We compute the average occurrence ratio of pervasive sensor events and GA-specific events and use it to split the GA-specific events. This event emphasizing helps to raise its frequency and prevent from being pruned in the pattern learning module. Since this step requires a GA-label of an event stream, it is processed only in the training phase.

\subsubsection{Noise Event Pair Removal}\label{filter} 
\begin{figure}[H]
  \centerline{\includegraphics[scale=0.2]{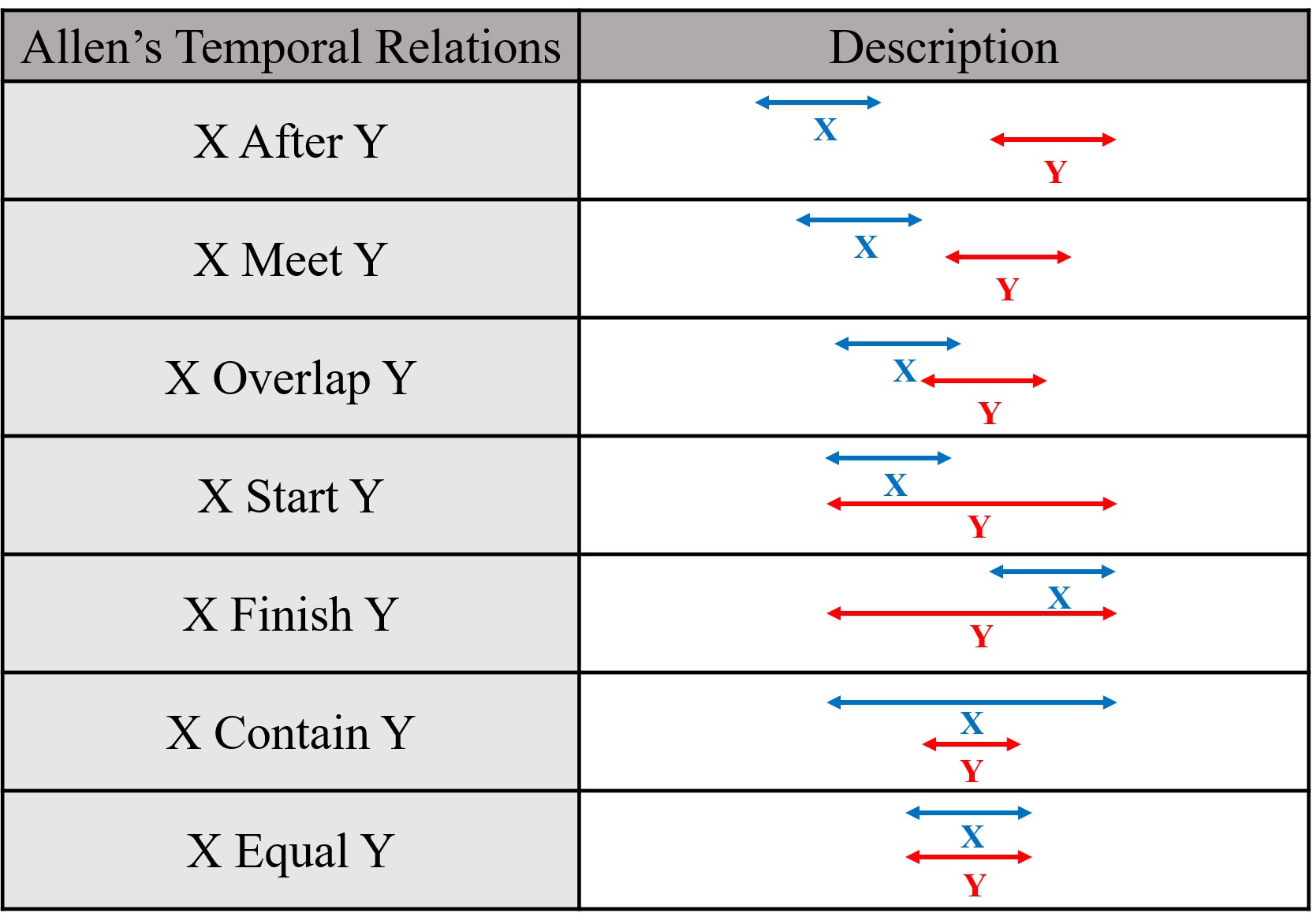}}
  \caption{Seven types of Allen's temporal relations}
  \label{fig5}
\end{figure}

Once emphasizing GA-specific events is done, the preprocessing module finds all possible pairs in a given event sequence to filter out non-causal noises. It compares the start time and the end time of a pair of events and assign the pair one of Allen's temporal relations \cite{allen1994} (Figure \ref{fig5}). Given a pair of events with \textit{After} relation, the module queries the knowledge base and \textit{event-causality-rule} and check if two events in the pair are causally related to each other. The following example describes how the rule verifies that  \textit{ProjectorOn} have causality with \textit{SoundLevel1} :  
 
\begin{table}[H]
\begin{center}
\begin{tabular}{ll}
$[$ event-causality-rule:\\[1.5 mm]
   ProjectorStatus - rd:type - ga:Context\\[1.5 mm]
   SoundLevel - rd:type - ga:Context\\[1.5 mm]
   ProjectorOn - rd:type - ga:Event\\[1.5 mm]

   SoundLevel2 - rd:type - ga:Event\\[1.5 mm]
   ProjectorOn - ga:isAbout - ProjectorStatus\\[1.5 mm]
   SoundLevel2 - ga:isAbout - SoundLevel\\[1.5 mm]
   ProjectorStatus - ga:relatedTo - SoundLevel\\[1.5 mm]
   $\longrightarrow$ ProjectorOn - ga:causes - SoundLevel2\\[1.5 mm]
$]$
\end{tabular}
\end{center}
\end{table}%

\subsubsection{Temporal Relation Extraction} :
If it does not have the \textit{causes}  relationship, a given pair of events is discarded. Event pairs passing the rule or having other six temporal relations are finally preserved. This step reduces noisy data among causal event pairs, thus, we can better extract the characteristics of each group activity.

We call the remaining relation as a \textit{causal relation} and represent it as follows:
\newline
\centerline{\{ RELATION (event\_x, event\_y) , start\_time, end\_time \}}
\newline
, where start\_time and end\_time are leveraged to determine temporal orders between causal relations. Eventually, a bundle of the determined Allen's relations represented by the first order logic (i.e. an episode) is temporarily stored in the episode storage as the training input dataset of the pattern learning module or is passed to the group activity recognition module as a test dataset.

\begin{algorithm}[t]
\SetAlgoNoLine
\KwIn{ Episodes of each GA $E$ and the min\_support threshold $T$ }
\KwOut{ A set of sequential patterns }
// About Lexicographic Tree:\

// Node $A$ is left brother of node $B$ if Node $A$ occurs before node $B$\

// Node $A$ is
father of node $B$ if sequence in $A$ node is a subsequence of sequence in node $B$.

$LexicographicTree$ <- frequent depth-1 nodes\;

//Itemset Extension

\For{each node $\alpha$ in $LexicographicTree$}{
\For{right brother nodes $\beta$ of node $\alpha$}{
      \If{concatenating node $\alpha$ and node $\beta$ >= $T$}{
       $LexicographicTree$<- concatenating node $\alpha$ and node $\beta$\;
      }
    }
}

//Sequence Extension

\For{each node $\alpha$ in $LexicographicTree$}{
\For{brother nodes $\beta$ of node $\alpha$ (or node $\alpha$ itself)}{
      \If{ concatenating node $\alpha$ and (node $\beta$ or node $\alpha$) >= $T$}{
       $LexicographicTree$<- concatenating node $\alpha$ and (node $\beta$\ or node $\alpha$);
      }
    }
}

Extract all nodes as patterns

\caption{FAST \cite{Salvemini2011} based pattern tree algorithm}
\label{alg:one}
\end{algorithm}

\begin{figure}[t]
  \centerline{\includegraphics[scale=0.4]{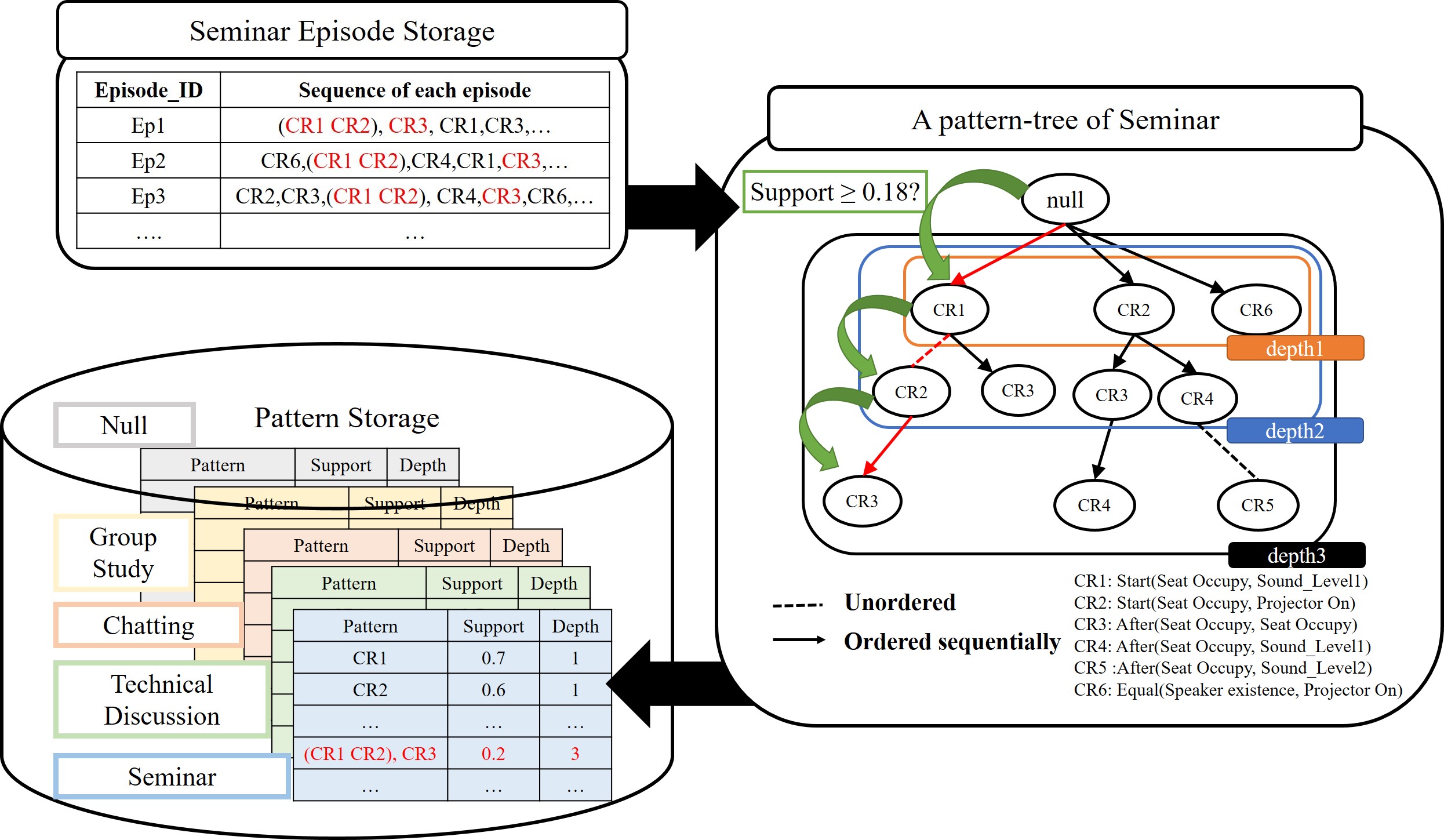}}
  \caption{Pattern-tree construction}
  \label{fig6}
\end{figure}
\pagebreak
\subsection{Pattern Learning Module} \label{pattern learning module}
To overcome the limitation of the association rule mining (that is, lower recognition accuracy than the graphical model), the sequential pattern learning module derives patterns which describe temporal orders between causal relations for richer representation. For that purpose, we leverage the pattern-tree algorithm \cite{Salvemini2011} which extracts frequent sequential patterns by the means of a growing tree structure as described in ALGORITHM \ref{alg:one}. 
Pattern growth algorithms support faster performance than other sequential pattern mining algorithms since they do not need to generate a large number of candidate sets. FAST algorithm has key features such as lexicographic tree, itemset extension, and sequence extension \cite{Salvemini2011}, which resuls in more efficient and scalable performance in terms of running time and memory usage than other pattern growth algorithms.

To input preprocessed GA episodes to a pattern-tree, we sort the causal relations contained in each episode by comparing their start\_time and end\_time in an ascending order. Then, for each GA, a pattern-tree is constructed by recursively adding causal relations with higher support values than a predefined support threshold. To avoid underfitting and overfitting, the threshold for each GA is set to a value that keeps the number of patterns existing in each GA similar.

Figure \ref{fig6} illustrates an example of the overall tree construction and patterns extraction procedure with an example GA, \textit {Seminar}. We choose the support threshold value, 0.18, for \textit{Seminar}, which is obtained from our experiments to get 2700 patterns (refer to Figure \ref{NumberOFPATTERNS}). When the construction starts, the episode storage containing a list of \textit {Seminar} episodes is loaded. Then, the pattern-tree thread looks up all episodes and adds causal relations with support values higher than 0.18 to the root node. For each causal relation in depth-1, the pattern-tree thread looks up all episodes again to find the following or co-occurrence causal relations with support values higher than 0.18 and adds them as depth-2 nodes. The figure shows causal relations with temporal orders represented by line arrows and causal relations that occur together represented by dotted arrows. The pattern-tree thread continues this look-up-and-add routine until no more patterns which has a higher support value than the threshold. When the tree construction is done, all possible subsequences of causal relations on a branch are stored to the pattern storage with their support and depth values. We express causal relations that occur together are expressed in parentheses and those with temporal precedence are represented by commas.

\subsection{Group Activity Recognition Module}

As pattern learning is done, the group activity recognition module can recognize a test GA event sequence using the pattern storage. The test sequence goes through the preprocessing module which extracts a sequence of causal relations as in the training phase. Then, causal relations in the sequence are sorted by comparing their start\_time and end\_time and the group activity recognition module accept them as test data as in the pattern learning module. The module examines whether given test data corresponds to the patterns of each GA, including causality.
To compute the likelihoods of stored GAs, we leverage the weight sum method of MCMC \cite{Antunes2001} that depends on both the number of matched patterns and their weights. A given test sequence is compared to each GA's learned patterns and the support values of matched patterns are added to the GA's likelihood as follows:

\begin{figure}[h]
\centering
  \centerline{\includegraphics[width=8.39cm,height=5.6cm]{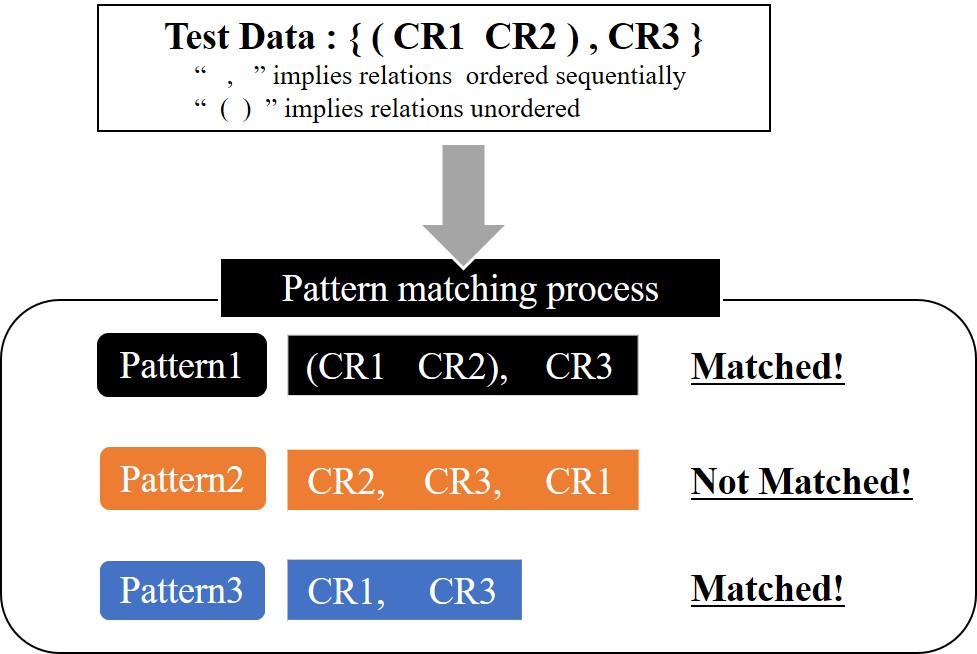}}
  \caption{A pattern matching mechanism in GAR module}
  \label{fig7}
\end{figure}

\begin{equation}
P(X = x) =\frac{1}{Z}e^{{\sum_i}{w_i}{n_i}{(x)}}\label{eqe}
\end{equation}
\newline
, where P(X = x) is the likelihood of a GA x, Z is a normalization factor. The normalization helps to avoid an error-prone situation in which all test sequences are recognized as a GA with high-support patterns in average. $i$ is the identifier of a pattern expressed as a number, $w_i$ is the support value of a matched pattern, and $n_i$  is a Boolean function, defined as follows:

\begin{equation}
{n_i}{(x)}=
\begin{cases}
    0,   & \text{if } i^{th} \text{ pattern of GA } x \text{ is not matched}\\
    1,             & \text{otherwise}
\end{cases}
\label{eq}
\end{equation}

Figure \ref{fig7} shows how the pattern matching process works. CR represents a causal relation and the relationships between causal relations are expressed as temporal orders (e.g. CR1 \& CR3, CR2 \& CR3) or co-occurrences (e.g. CR1 \& CR2). Given test data is matched to Pattern1 and Pattern3 because CR3 appears later than CR1 or CR2. However, in the case of Pattern2, the test data can not be matched since CR3 does not occur before CR1. The support values of Pattern1 and Pattern3 are applied to equation \ref{eqe} as weights and we can obtain the probability of the test data.

After all probabilities for each GA are calculated using equation \ref{eqe}, the most probable GA is recognized as a result.

\section{experiment results}
\subsection{Experiment Overview}
We conduct experiments in a testbed in our university, a meeting room where we install various sensors and smart objects for capturing everyday usages. We intend to evaluate the proposed scheme in terms of group activity recognition (GAR) accuracy, runtime performance, and 
robustness in a real environment where data is missing or false values are generated due to errors or malfunctions of some pervasive sensors or actuators, compared with three existing schemes, Association Rule Mining \cite{lotfi2016}, CRP-based ITBN \cite{liu2016} and HMM \cite{Eddy1996}. 
In the following subsections, we first explain the hardware and software configurations for our experiments. Then, we describe the experiment results in detail in terms of three metrics above. For accuracy evaluation, we analyze how the proposed scheme supports both causality and structural flexibility impact accuracy in comparison to the existing schemes. We also measure the runtime performance of each scheme and show that the proposed scheme performs with less computation overhead even if the number of event episodes increases, required for better accuracy. As to prove tolerance to missing or false data, we show that the recognition accuracy of the proposed scheme degrades much less than the graphical model-based scheme. To verify the scalability of our scheme in other smart spaces, we also evaluate the accuracy of the proposed scheme and the existing ones with CASAS datasets \cite{CASAS,CASAS2}. Finally, we discuss the remaining issues that affect the recognition accuracy and further improvement points.

\begin{figure*}[t]
\centering
  \centerline{\includegraphics[scale=0.55]{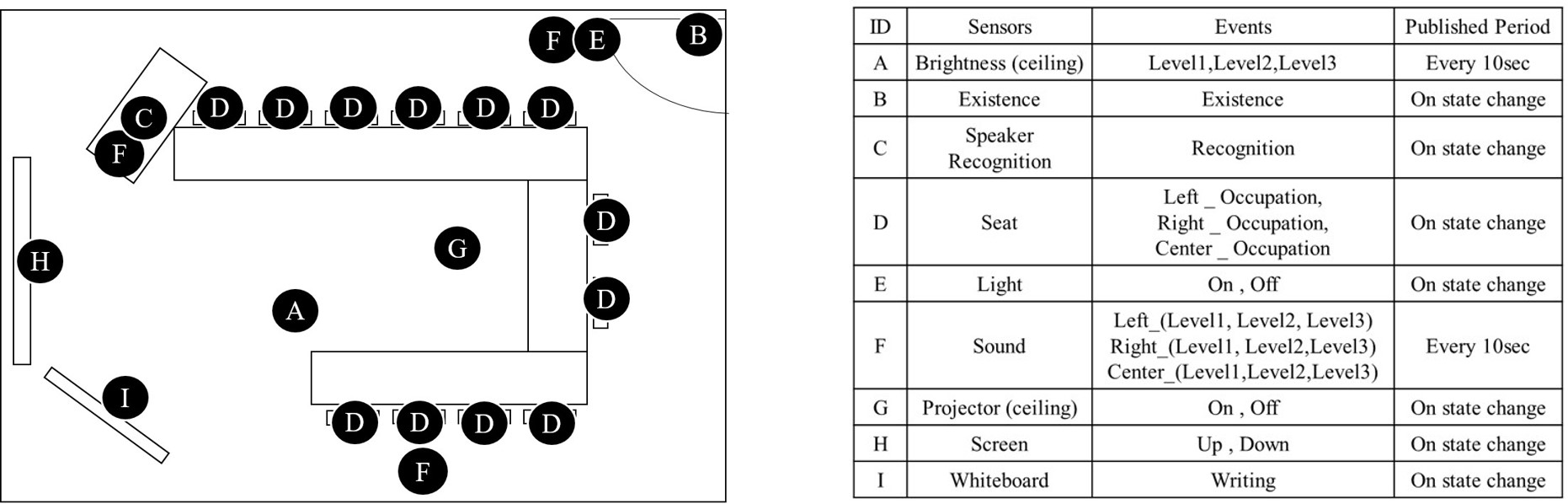}}
  \centerline{(a) Smart objects installation in an IoT testbed$$\qquad\qquad\qquad\quad$$(b) Types of events generated by smart objects}
  \caption{ Pervasive sensor space testbed construction}
  \label{fig8}
\end{figure*}

\begin{figure}[]
\centerline{\includegraphics[scale=0.45]{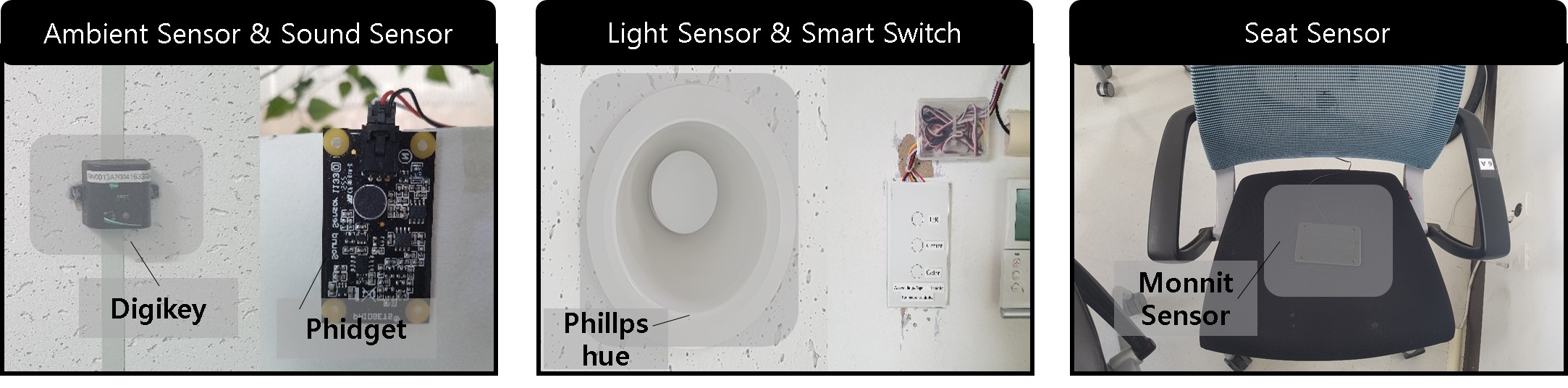}}
  \caption{ Deployed smart objects in an IoT testbed}
  \label{fig9}
\end{figure}

\subsection{Experiment Setup}
 We conduct our experiments on Intel Xeon CPU E5-2620 v3 (2.4GHz), 16GB RAM, and Window 10 pro OS.
 For a comparative experiment, we implement our scheme and the three schemes mentioned above in Java. As inputs, the tuples of events (start time, end time, event type) are given to our scheme, the CRP-based ITBN scheme, and the association rule mining scheme while the timestamps of events are given to the HMM scheme. For preprocessing modules, Allen's all temporal relations are applied to our scheme and the CRP-based ITBN scheme, while six temporal relations without `After' relation are applied to the association rule mining scheme. To implement each pattern learning module, we leverage well-known open source java libraries as follows: SPMF \cite{SPMF} for sequential mining of our scheme and the association rule mining scheme, datumbox \cite{datubox} for the CRP-based ITBN scheme, and wiigee \cite{wii} for the HMM scheme. For activity recognition modules, we implement the weighted sum method for our scheme and the association rule mining scheme. We leverage the inference packages in \cite{datubox, wii}.
 
We install twenty-three smart objects units in our testbed which produce nine different types of events. Figure \ref{fig8}-(a) shows the installation configuration of smart objects in the testbed and Figure \ref{fig8}-(b) describes the detailed explanation of sensors and their publishing events. Figure \ref{fig9} shows how pervasive sensors and actuators are implemented using Raspberry Pi 3 and commercial sensor frameworks (e.g. Phidgets \cite{phidget} and DigiKey \cite{digikey})

\subsection{Testbed Dataset Description} 
Existing public datasets \cite{CASAS,opportunity} for group activity recognition schemes have several limitations. First, they are generated by controlled experiments in which well-defined instructions are given to participants. It implies that they may not fully take into consideration situations that can occur in an actual system (e.g. Situations involving noise). Second, they do not involve enough pervasive sensors to keep track of various patterns in group activities. For example, CASAS group activity dataset \cite{CASAS} involves only IR motion sensors and cabinet sensors as pervasive ones and Opportunity dataset \cite{opportunity} is biased towards wearable sensors rather than pervasive sensors. 
\begin{table*}[t]
\centering
\caption{DESCRIPTION OF TARGET GROUP ACTIVITIES}
\begin{tabular}{|m{1.6cm}|m{9.4cm}|c|c|} 
 \hline
\thead{GA\\ \textbf{Category}} & \makecell{\textbf{Description}}& \textbf{\# of people} & \thead{\textbf{Average} \\ \textbf{\# of events}} \\  
 \hline
 \makecell{Chatting} &  - A pair of people sit down nearby and have a casual conversation
 & 2 $\sim$ 3 & \makecell{19\\(A : 3, P : 16)}\\ 
 \hline
\makecell{ Technical \\ Discussion} &  \makecell{ - Using a projector to display discussion topics, \\ many people discuss about them}
 &  4 $\leq$ & \makecell{49\\(A : 7 , P : 42)}\\ 
 \hline
 \makecell{ Seminar} & \makecell{- One or more speakers make presentations,\\ others discuss topics of the presentations}
 &  4 $\leq$ & \makecell{47\\(A : 12 , P : 35)}\\ 
 \hline
 \makecell{Group \\ Study} & \makecell{ - Some people study together for a long time }
 &  2 $\sim$ 3  & \makecell{61\\(A : 6 , P : 55)}\\ 
 \hline
 \makecell{ Null} & \makecell{ - No one exists}
 &  0 & \makecell{31\\(A : 0 , P : 31)}\\ 
 \hline
 
\end{tabular}

\label{table:1}
  \bigskip\centering
 * \emph{A :} Actuators , \emph{P :} Pervasive Sensors
\end{table*}
To overcome such limitations and collect richer dynamics of group activity patterns, we collect 3,466,339 IoT sensor records for 14 months from our testbed where students and faculty members conduct actual group activities. We translate the collected raw data into event level according to changes of raw data values. As shown in Figure \ref{fig8}, each sensor publishes raw data when the sensor state changes or according to a fixed interval and convert to event data. We choose five dominant GAs conducted by users which are defined in TABLE \ref{table:1} and we use 428 episodes of those GAs in total for learning and recognizing in experiments.

\subsection{Experiment Results: Accuracy with our testbed dataset}\label{Experiment Results: Accuracy with our testbed dataset}

\subsubsection{Optimal number of rules}\label{oprimalnumber}

\begin{figure}[H]
\centerline{\includegraphics[scale=0.7]{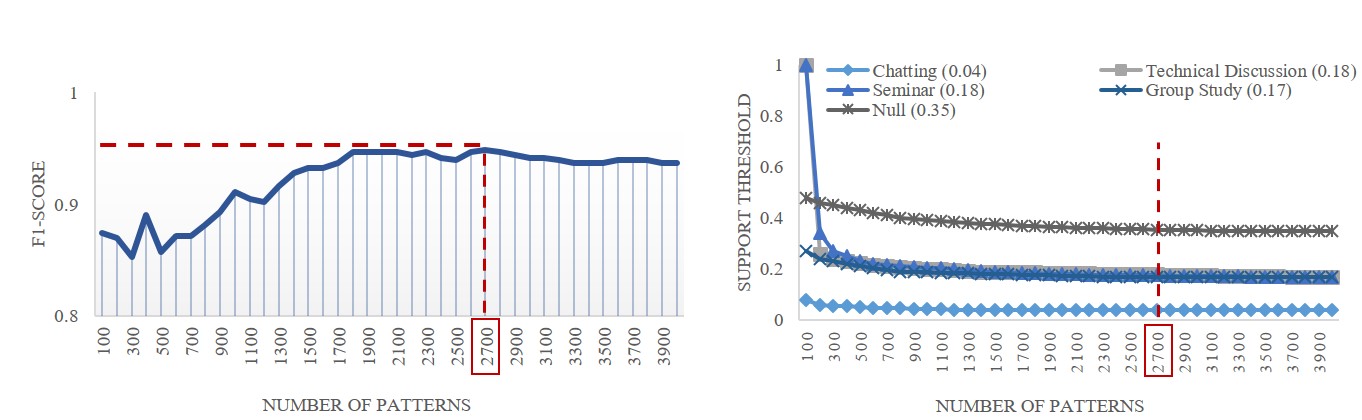}}
 \centerline{\qquad(a) F1-score $\qquad\qquad\qquad\qquad\qquad\qquad\qquad$(b) Threshold \qquad}
  \caption{F1-score and threshold according to the number of patterns}
  \label{NumberOFPATTERNS}
\end{figure}

As mentioned in Section \ref{pattern learning module}, we determine the support threshold of each GA such that all GAs will have the same number of learned patterns. Finding the optimal number of patterns is important to avoid underfitting or overfitting problems. We conduct an accuracy experiment by varying the number of patterns from 100 to 4000 as shown in Figure \ref{NumberOFPATTERNS}-(a). Starting with a F1-score 0.87 in 100 patterns, the accuracy increases slightly until the number of patterns is 1700. When the number of patterns exceeds 1700, F1-score accuracy is converged between 0.94 and 0.95 until 2700 patterns. However, after then, overfitting problems occur and the accuracy decreases little by little as the number of patterns increases. The accuracy is the best with 2700 patterns and we finally set 2700 as the optimal number of patterns in our experiments. Figure \ref{NumberOFPATTERNS}-(b) shows the correlation between the threshold value and the number of patterns for each GA, which explains why we choose a different threshold value for each GA. The number in parentheses following a GA name in Figure \ref{NumberOFPATTERNS}-(b) represents the threshold value for each GA.

\begin{figure}[!b]
\centerline{\includegraphics[scale=0.9]{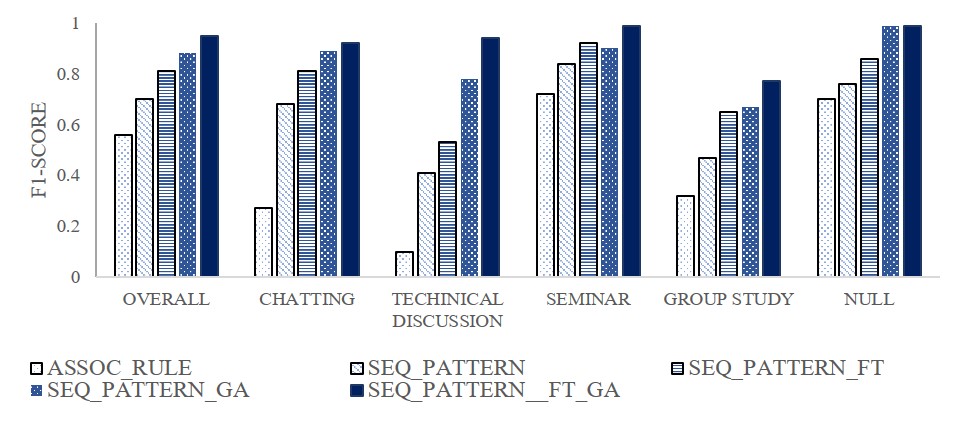}}
  \caption{Effects of causality awareness preserving mechanisms}
  \label{EffectsCausalities}
\end{figure}

\subsubsection{Effects of causality awareness preserving mechanisms}

We evaluate the effectiveness of three causality awareness preserving mechanisms in the proposed scheme (pattern learning in Section \ref{pattern learning module}, noise event pair removal in Section \ref{filter}, and GA-specific event preservation in Section \ref{ga-specific}) in terms of accuracy by comparing five different test cases: \textit{ASSOC\_RULE}, \textit{SEQ\_PATTERN}, \textit{SEQ\_PATTERN\_FT}, \textit{SEQ\_PATTERN\_GA}, and \textit{SEQ\_PATTERN\_FT\_GA}.
In \textit{ASSOC\_RULE} case, we evaluate the accuracy when frequent patterns with Allen's temporal relations are extracted by association rule mining which does not preserve causality. 
\textit{SEQ\_PATTERN} case is to evaluate the accuracy of the pattern learning module which runs without noise event pair removal and GA-specific event preservation. 
To evaluate the accuracy of the causality-related preprocessing mechanisms, we design \textit{SEQ\_PATTERN\_FT} and \textit{SEQ\_PATTERN\_GA} cases where the noise event pair removal and GA-specific event preservation are applied to the pattern learning module, respectively.
In \textit{SEQ\_PATTERN\_FT\_GA} case, all proposed mechanisms are included and causality awareness is the most well preserved among all cases. The comparison results are shown in Figure \ref{EffectsCausalities}.

\textit{SEQ\_PATTERN}, \textit{SEQ\_PATTERN\_FT}, \textit{SEQ\_PATTERN\_GA}, and \textit{SEQ\_PATTERN\_FT\_GA} improve accuracy by 14\%, 25\%, 32\%, and 39\%, respectively, compared to \textit{ASSOC\_RULE}. Accuracy is improved in \textit{SEQ\_PATTERN} for all GAs since causality makes learned patterns more descriptive by ordering temporal relations and eventually distinguishes GAs better. 
Especially, the accuracy of \textit{Chatting}, \textit{Technical Discussion} and \textit{Group Study} is drastically improved since users conduct similar activities yet in different orders when performing those GAs.

In \textit{SEQ\_PATTERN\_FT}, the proposed filtering mechanism removes the pairs of events that do not affect each other (i.e. no causality) and frequently appear such as `\textit{After(sound, sound)}'. 
Such event pairs are not informative but disturb GA recognition because they can overwhelm other important event pairs. The evaluation results show that improvement by filtering out non-causal event is achieved in all GAs (11\% improvement in average compared to \textit{SEQ\_PATTERN}). 
In \textit{SEQ\_PATTERN\_GA}, GA-specific events are emphasized (i.e. scale-up) such that they can be recognized as important patterns. The accuracy of \textit{Technical Discussion} and \textit{NULL} considerably increase since the GA-specific event preservation mechanism enables `Project ON' or `Light OFF' events to be notable in the learned patterns.(18\% improvement in average compared to \textit{SEQ\_PATTERN}).  \textit{SEQ\_PATTERN\_FT\_GA} takes all advantages of three mechanisms and draws significant improvement on GA recognition accuracy.

\begin{table*}[!b]
  \centering
  \caption{RECOGNITION ACCURACY COMPARISON RESULTS}
\begin{minipage}{\columnwidth}
\begin{center}
\begin{tabular}{|m{2cm}|m{0.7cm}|c|c|m{1cm}|c|c|m{0.5cm}|c|c|m{0.5cm}|c|c|}
\hline
\textbf{\centerline{Group}}&\multicolumn{3}{|c|}{\textbf{Proposed Scheme}}&\multicolumn{3}{|c|}{\textbf{Association Rule Mining}}&\multicolumn{3}{|c|}{\textbf{CRP-based ITBN}}&\multicolumn{3}{|c|}{\textbf{HMM}}  \\
\cline{2-13} 
\textbf{\centerline{Activities}} & \textbf{\textit{\centerline{S}}}& \textbf{\textit{R}}& \textbf{\textit{F1}} &\textbf{\textit{\centerline{S}}}& \textbf{\textit{R}}& \textbf{\textit{F1}}
&\textbf{\textit{\centerline{S}}}& \textbf{\textit{R}}& \textbf{\textit{F1}}
&\textbf{\textit{\centerline{S}}}& \textbf{\textit{R}}& \textbf{\textit{F1}}\\
\hline
\centering{Chatting} 
& \centering{0.96}& {0.94} & {0.92} 
& \centering{0.86} & {0.82} & {0.75}
&\centering{0.97} & {0.65}& {0.75} 
&\centering{0.93}& 0.56 &  0.65\\
\hline
\centering{Technical Discussion} 
& \centering{0.99}& {0.94} & {0.94} 
& \centering{0.95} & {0.52}& {0.55} 
&\centering{0.95} & {0.67} & {0.66}
&\centering{0.94}& {0.62} &  {0.60}\\
\hline
\centering{Seminar} 
&\centering{0.99}& {0.99} & {0.99} 
& \centering{0.97} &{0.73}&{0.82}
&\centering{0.86} & {0.79} & {0.75} 
&\centering{0.91}& {0.76} & {0.78}\\
\hline
\centering{Group Study} 
&\centering{0.98}& {0.73} & {0.77} 
& \centering{0.96} & {0.38}& {0.42}
&\centering{0.95} & {0.49} & {0.48}
&\centering{0.84}& {0.67} & {0.41}\\
\hline
\centering{Null Task} 
&\centering{0.99}& {0.99} & {0.99} 
& \centering{0.93} & {0.97} & {0.86}
&\centering{0.96} & {0.99} & {0.93} 
&\centering{0.95}& {0.61} &  {0.67}\\
\hline
\hline
\centering{\textbf{Average}} 
&\centering{\textbf{0.99}}& \textbf{0.95} & \textbf{0.95}
&\centering{\textbf{0.94}} & \centering{\textbf{0.74}} & \centering{\textbf{0.74}}
&\centering{\textbf{0.94}} & \centering{\textbf{0.76}} & \centering{\textbf{0.76}}
&\centering{\textbf{0.91}}& \textbf{0.65} &  \textbf{0.65}\\
\hline
\end{tabular}
\end{center}
  \label{table:2}
  \bigskip\centering
 * \emph{S :} Specificity, \emph{R :} Recall, \emph{F1 :} F1-score
\end{minipage}
\end{table*}

\subsubsection{Comparison with other schemes}
To assure that the proposed scheme recognizes group activities better than the existing ones, we conduct a leave-one-out cross validation to learn as many types of activity patterns as possible because of group activities' complexity. We use specificity, recall, precision and F1-score as recognition accuracy measures. 
As shown in TABLE \ref{table:2}, the proposed scheme performs better than the existing ones, Association Rule Mining \cite{lotfi2016}, CRP-based ITBN \cite{liu2016}, and Hidden Markov Model (HMM) \cite{Eddy1996}, in terms of all accuracy measures. The proposed scheme improves the F1-score by 21\%, 19\%, and 30\% compared to association rule mining scheme, CRP-based ITBN scheme, and HMM scheme in average, respectively. Recall and F1-score have the same value in all schemes because the same equation is derived when $\Sigma$ assigns to each of element (e.g. $TP$ = ${\sum_i}{TP_i}$). In specificity measure, the proposed scheme performs 5\%, 5\%, and 8\%  better than the other ones.

\begin{figure}[]
\centerline{\includegraphics[scale=0.8]{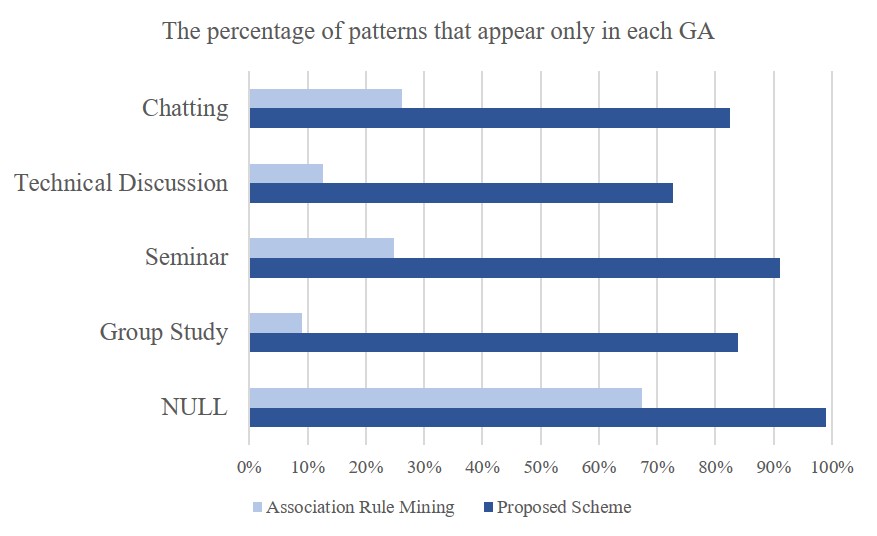}}
\caption{Ratio of each GA's unique patterns}
\label{fig10}
\end{figure}

The proposed scheme aims to improve recognition accuracy by preserving potential temporal event orders. We validate the impact of adding causality for better recognition rates, compared to the existing pattern-based models. The proposed scheme shows a significantly higher F1-score in all GAs than the association rule mining scheme. The proposed scheme shows much higher scores in specificity value for \textit {Chatting} (0.96) and in recall rate for \textit {Technical Discussion} (0.94) , \textit{Seminar} (0.99) and \textit {Group Study} (0.73) than the existing scheme. This implies that many test event sequences of \textit {Technical Discussion}, \textit{Seminar} and \textit {Group Study} are less recognized as \textit{Chatting} in the proposed scheme compared to related work. Our scheme preserves causality so that the patterns of \textit {Chatting} are distinct from patterns of other GAs.
Also, our scheme less recognize test event sequences of \textit {Seminar} as \textit {Technical Discussion}. According to the GA definitions in TABLE \ref{table:1}, \textit {Seminar} is different from \textit {Technical Discussion} in that the former has a main speaker and a projector (i.e. GA-specific event) is used for presentation. Preserving an order between events makes it possible to differentiate similar situations of those GAs like speakers taking turns in order (\textit {Seminar}) from others where discussion participants are joining and leaving the room during discussion with projector (\textit {Technical Discussion}). The proposed scheme extracts more GA-unique patterns for five activities used in our experiment as shown in Figure \ref{fig10} and it leads to a high recognition rate than the association rule mining scheme.  

Note that the proposed scheme recognizes all GAs better than the CRP-based ITBN scheme which is the best accuracy method among graphical models. It is because our test data include event sequences that do not exactly match with the ones obtained from the training data. 
Especially, the specificity value for \textit {Seminar} (0.99) and the recall rates for all GAs of the proposed scheme are higher than those of the CRP-based ITBN scheme. This corresponds to why the proposed scheme less mistakes \textit {Seminar} test event sequences for other GAs than the CRP-based ITBN scheme. The detailed explanation on this will be given in the following Section \ref{tolerance}.
As described in the Section \ref{filter}, the proposed scheme handles concurrent events using Allen's temporal model. It allows the proposed scheme to perform better than Hidden Markov Model (HMM) in all accuracy measure scores.

\subsection{Experiment Results: Runtime Performance}

\begin{figure}[H]
\centerline{\includegraphics[scale=0.4]{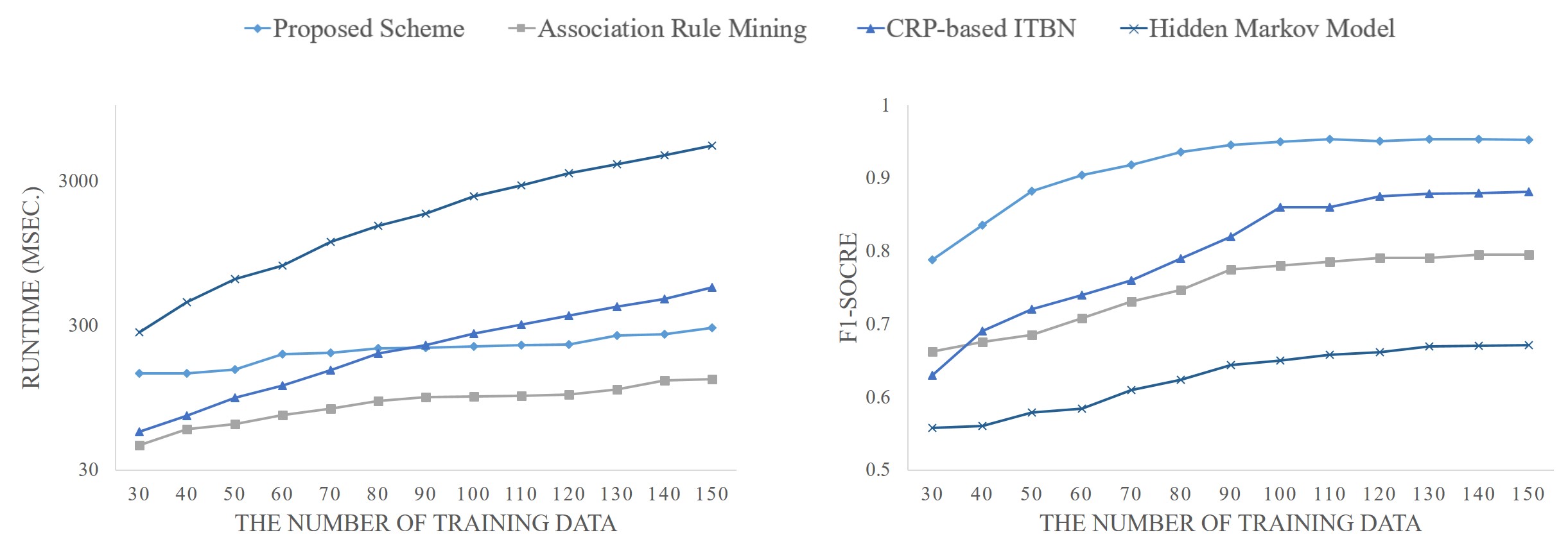}}
 \centerline{\qquad\quad(a) A comparison of runtime$\qquad\qquad\qquad\qquad\qquad\quad\quad$(b) A comparison of accuracy \qquad}
\caption{Comparisons of runtime and accuracy  form 30 to 150 training data}
\label{fig11}
\end{figure}

We measure the run-time performance of the proposed scheme and the existing ones as the number of GA episodes increases. Since any GA is hardly trained with less than 30 episodes, we increase the number of episodes from 30 to 150. We randomly select required number of episodes from 150 episodes and repeat this process multiple times. The mean values of the results are shown in Figure \ref{fig11}. As shown in Figure \ref{fig11}-(a), the proposed scheme and the association rule mining scheme consumes a relatively constant amount of computation while  the execution time of CRP-based ITBN scheme and HMM scheme continuously increases. The proposed scheme takes a little more time than the association rule mining scheme since it involves causality handling. 

Figure \ref{fig11}-(b) shows that all schemes reach maximum accuracy as the data size is larger than 100. This convergence implies that the accuracy results from Table \ref{table:2} hold even with a larger dataset. We also observe that the proposed scheme reaches the accuracy convergence point earlier than other schemes as shown in Figure \ref{fig11}. We can obtain an accuracy of 0.78 with only 30 training data and can get above 0.9 F1-score  when training with more than 60 episodes. This  shows that the proposed scheme can guarantee better accuracy than the graphical models with less run-time overhead. 

\subsection{Missing and False Data Tolerance: vs. CRP-based ITBN scheme}\label{tolerance} 
\begin{figure}[]
\centerline{\includegraphics[scale=0.55]{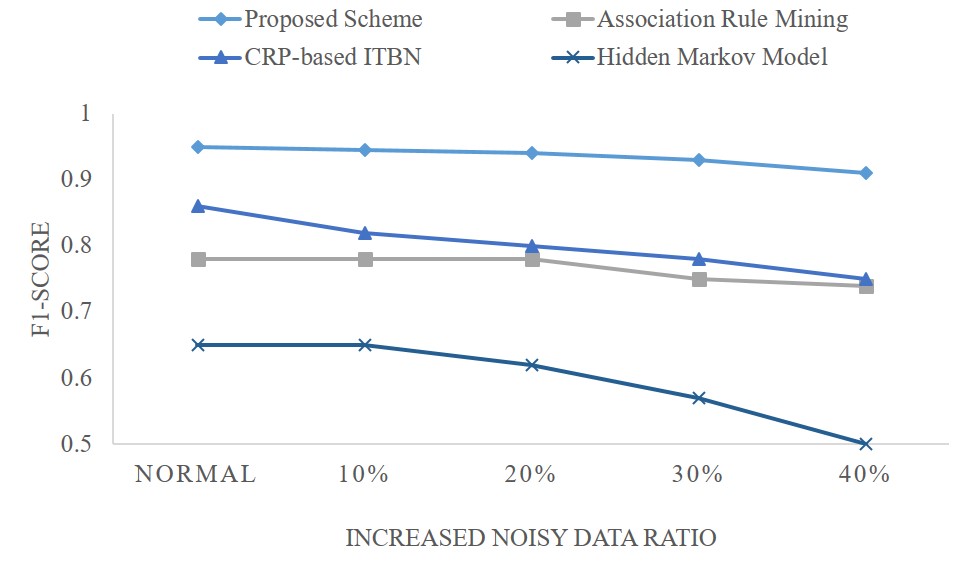}}
\caption{A noise analysis result by increasing the number of noisy data}
\label{fig12}
\end{figure}

GAR solutions for a pervasive sensor space should be robust and cope with the dynamic nature of pervasive sensors data such as missing or false values. The proposed scheme's key improvement over the graphical model is its flexible representation and we leverage this flexibility to enhance the tolerance of missing or false data. 
We refer to an episode that includes missing or false event values as a noisy episode. The observation in our testbed for six months confirms that such noisy episodes often happen in a real environment.
We increase the noisy episode ratio in the training dataset to 10\%, 20\%, 30\% and 40\% and compare the performance degradation tendency of the proposed scheme and the CRP-based ITBN scheme. Figure \ref{fig12} shows that F1-scores decrease as the noise data ratio (i.e. the percentage of episodes with random missing or false data among all episodes) increases. In average, the F1-score degradation ratio of the proposed scheme is 0.015 per 10\% noise data, while that of the CRP-based ITBN scheme is 0.0425. This result indicates that the proposed scheme is more robust than the graphical model by almost three times. 

The faster performance degradation of the CRP-based ITBN scheme is caused by its structural characteristics. 
When some false events are injected or some events are missed to a training event sequence, it disrupts the state transitions which are supposed to be extracted as frequent patterns.
The more noisy episodes are added, the more disruptions happen. On the other hand, the proposed scheme performs the flexible computations of temporal and causal relations between event pairs with little disruptions even if false events are added or some events are missed to the training event sequences. 

\begin{table*}[!b]
  \centering
  \caption{RECOGNITION ACCURACY COMPARISON RESULTS (CASAS group activity dataset)}
\begin{minipage}{\columnwidth}
\begin{center}
\begin{tabular}{|m{2cm}|m{0.7cm}|c|c|m{1cm}|c|c|m{0.5cm}|c|c|}
\hline
\textbf{\centerline{Group}}&\multicolumn{3}{|c|}{\textbf{Proposed Scheme}}&\multicolumn{3}{|c|}{\textbf{Association Rule Mining}}&\multicolumn{3}{|c|}{\textbf{CRP-based ITBN}}
\\
\cline{2-10} 
\textbf{\centerline{Activities}} & \textbf{\textit{\centerline{S}}}& \textbf{\textit{R}}& \textbf{\textit{F1}} &\textbf{\textit{\centerline{S}}}& \textbf{\textit{R}}& \textbf{\textit{F1}}
&\textbf{\textit{\centerline{S}}}& \textbf{\textit{R}}& \textbf{\textit{F1}}\\
\hline
\centering{Move furniture} 
& \centering{0.96}& 0.99 & 0.95 
& \centering{0.97} & 0.99 & 0.96
&\centering{0.95} & 0.69 & 0.75 \\

\hline
\centering{Play a game} 
& \centering{0.99}& 0.92 & 0.94 
& \centering{0.96} & 0.96 & 0.93
&\centering{0.99} & 0.92 & 0.96 \\

\hline
\centering{Prepare for dinner} 
&\centering{0.92}& 0.92 & 0.86 
& \centering{0.91} & 0.41 & 0.47
&\centering{0.99} & 0.46 & 0.63 \\

\hline
\centering{Pack a picnic} 
&\centering{0.99}& 0.72 & 0.82 
& \centering{0.84} & 0.60 & 0.58
&\centering{0.74} & 0.99 & 0.71 \\
\hline
\hline
\centering{\textbf{Average}} 
&\centering{\textbf{0.96}}& \textbf{0.89} & \textbf{0.89}
& \centering{\textbf{0.92}} & \textbf{0.76} & \textbf{0.76}
&\centering{\textbf{0.92}} & \textbf{0.77} & \textbf{0.77} \\

\hline
\end{tabular}
\end{center}
  \label{table:3}
  \bigskip\centering
 * \emph{S :} Specificity, \emph{R :} Recall, \emph{F1 :} F1-score
\end{minipage}
\end{table*}

\subsection{Experiment Results: Accuracy with CASAS datasets}
To validate the scalability of the proposed scheme in terms of applicability to various environments, we conduct an additional experiment using the CASAS group activity dataset \cite{CASAS} after removing users ID. 
The CASAS group activity dataset contains 26 episodes of four group activities: \textit{Move furniture}, \textit {Play a game}, \textit {Prepare for dinner} and \textit {Pack a picnic}. It was collected from a smart home testbed where 51 motion sensors and 15 cabinet sensors were used and two residents performed collaborative tasks together. 
The sensor configuration like this implies that the recognition scheme should exploit event sequences representing users' movement paths for identifying activity patterns. 

In this experiment, the same practice for measuring the accuracy in Section \ref{Experiment Results: Accuracy with our testbed dataset} is applied to the CASAS group activity dataset. Since \textit{Pack a picnic} is too complicated to derive as a basic HMM model, we compare the results of ours, association rule mining scheme, and CRP-based ITBN scheme without HMM scheme as shown in TABLE \ref{table:3}. In summary, our scheme outperforms the other two schemes with respect to all accuracy metrics. In addition, since the dataset consists of only two sensors (motion and cabinet) with limited expression power, overall F1-scores are lower than those in Section \ref{Experiment Results: Accuracy with our testbed dataset}. All three schemes recognize \textit{Play a game} with high accuracy because it consists of different user movement paths compared to other group activities.

We first observe that the proposed scheme and the association rule mining scheme often recognize \textit {Pack a picnic} as \textit {Prepare for dinner}, which degrades corresponding recall rates down to 0.72 and 0.60, respectively. This is because incorrectly recognized episodes contain the common patterns representing both group activities, but \textit{Prepare for dinner} has more GA-unique patterns. Since \textit {Pack a picnic} contains much more common patterns than the other GAs, the accuracy of \textit {Pack a picnic} is decreased.
On the other hand, the CRP-based ITBN scheme shows a higher recall rate on \textit{Pack a picnic} (0.99). This is because CRP-based ITBN scheme constructs the most various ITBN models for \textit{Pack a picnic} and it makes \textit{Pack a picnic} well recognized, not mistaken by other group activities. The proposed scheme recognizes \textit{Prepare for dinner} much better (0.92 recall rate) than the other schemes (0.41 and 0.46). This is because the proposed scheme preserves potential temporal orders (i.e. causality) between events within patterns and it helps to distinguish user movement sequences better. Hence, the learned user movement patterns for \textit {Prepare for dinner} are differentiated from those for \textit {Pack a picnic}.

We also find that the recall rates of the CRP-based ITBN scheme on \textit{Move furniture} (0.69) and \textit{Prepare for dinner} (0.46) are much lower than the proposed scheme. This is because the scheme recognizes \textit{Move furniture} and \textit{Prepare for dinner} as \textit{Pack a picnic} since \textit{Pack a picnic} contains users' movement paths which also belong to \textit{Move furniture} and \textit{Prepare for dinner}. If a test event sequence of \textit{Move furniture} or \textit{Prepare for dinner} has different users' movement paths compared to the learned models, it tends to be recognized as \textit{Pack a picnic} because a learned model of \textit{Pack a picnic} contains such extra user movement paths. Moreover, as the number of training sets increases, even though the CRP-based ITBN models of \textit{Move furniture} or \textit{Prepare for dinner} include the extra user movement paths, they may be contained in the \textit{Pack a picnic} CRP-based ITBN model, too. Therefore, \textit{Move furniture} and \textit{Prepare for dinner} are prone to be recognized as \textit{Pack a picnic}. On the other hand, our scheme assigns different weights to each GA even for the same pattern. This feature enables \textit {Move furniture} and \textit {Prepare for dinner} to be distinguished better from \textit{Pack a picnic} if the extra user movement paths are more important in those two GAs. 

%

To verify if the proposed scheme recognizes single user-driven complex activities with high accuracy as well, we conduct one more experiment with the CASAS single user complex activity dataset \cite{CASAS2}. In this CASAS dataset, each of twenty one single users independently performed eight activities in a random order: \textit{fill medication dispenser}, \textit{wash dvd}, \textit{water plants}, \textit{answer the phone}, \textit{prepare birthday card}, \textit{prepare soup}, \textit{clean} and \textit{choose outfit}. 51 motion sensors, 8 item sensors (attached to specific objects), 12 cabinet sensors, 2 water sensors, a burner sensor, a phone sensor and 3 temperature sensors records are collected from the testbed. Compared to CASAS group activity dataset \cite{CASAS}, it describes specific activities better and the proposed scheme produces higher average F1-scores (0.9) than that of another related work \cite{riboni2016} (0.81). This result validates that the proposed scheme accurately recognizes complex activities driven by a single user.

\section{Discussions and Future Works}

\subsection{Improving Recognition Accuracy of Group Activities (GAs)  using Activity Hierarchy}
As shown in in Table \ref{table:2}, the proposed scheme yields a lower specificity  value (0.96) for \textit{Chatting} and a lower recall value (0.73) for \textit{Group Study} than the other GAs. This is because some of \textit{Group Study} episodes are mistakenly recognized as \textit{Chatting} ones.  
That is, some participants in a \textit{Group Study} session may focus more on chatting with other participants than just studying. This makes the proposed scheme incorrectly interpret the event sequences generated by such \textit{Chatting} as important patterns for \textit{Group Study} and thus the specificity of \textit{Chatting} decreases. One of solutions to fix this problem could be to define a hierarchy between GAs of inclusion relationship and infer such GAs from their differences but commonalities.

\subsection{Shallow vs. Deep Classifiers for Activity Recognitions}
While the schemes including ours mentioned in this paper are based on shallow classifiers, deep learning architectures have been recently presented for training public large datasets in the Human Activity Recognition (HAR) domain. Krishna et al. \cite{krishna2018lstm} present an LSTM network to train correlations between user activities and their durations by using a public dataset, American Time Use Survey (ATUS). The trained network predicts a sequence of next user activities with their durations. Guan et al. \cite{guan2017ensembles} present ensemble of a set of LSTMs to learn user activity patterns from widely used wearable sensor datasets by community researchers. Ronao et al. \cite{ronao2016human} use one-dimensional time-series signals from smartphones to train a Convolutional Neural Network (CNN) for user activity pattern recognition. To extend this one-dimensional data analysis to multi-modal data sources, Radu et al. \cite{radu2018multimodal} and Hossain et al. \cite{hossain2018} recently present a case study to apply deep learning techniques which 
features out of multi-modal sensors and perform 
a human activity recognition task.

However, to the best of our knowledge, no deep-learning architecture has been proposed for GAR in pervasive sensor environments. This is because while various well-structured and large-scale datasets collected from wearable devices or online social media are available for deep learning, none has been released for GAR due to privacy protection, difficulty of testbed installations, and so on. 
For instance, we collected pervasive sensor data streams for three different group activities for six months but only about 100 episodes can be leveraged for training group activity patterns due to privacy protection and noise filtering. 
%
By using these non-large datasets, most existing approaches and the proposed scheme show higher than or near 0.9 F1 scores even with shallow classifiers. Of course, this high recognition accuracy may come from the excellence of the schemes. However, it is also possible that the embedded variations of user behaviors in the given datasets are not rich in terms of group activity types and the number of corresponding event sequences. Even with the decent accuracy results, we cannot guarantee the dataset embraces the whole potential group activity dynamics. This is why the release of a large dataset collected for GAR in pervasive sensor spaces should be a great contribution for the community. Once this challenge is resolved and a large-scale dataset enough to encompass more group activity dynamics is widely used, many outperforming applications of deep learning techniques will advance GAR domain a lot. This is one of our currently ongoing efforts.

\subsection{Future Works}
As future work, we plan to advance a parameter optimization mechanism to incorporate different large datasets from different spaces flexibly. The proposed scheme has a hybrid architecture of heuristics and a data mining technique and their balance affects the recognition accuracy. The parameter $N$ to duplicate causally related event pairs is devised for the balancing purpose and a value $10$ is assigned considering the ratio between GA-specific events and others. However, people conduct different group activities in different spaces, which requires the parameter $N$ should adapt accordingly. It is highly possible that recognition accuracy increases with a higher $N$ value when many group activities can be characterized by specific events and their states are captured as sensor events. On the other hand, when there is no GA-specific event characterizing a group activity, the $N$ value should be lower so that its impact decreases. When the dataset is large, it becomes more complicated to find the optimal balancing parameter. It requires too much overhead to manually define GA-specific events and compute their ratio to other events by looking up the whole dataset. Since domain experts or application engineers should be involved with the GA-specific event definition procedure, it is difficult to automate the optimal parameter detection. Therefore, to guarantee the current outperforming accuracy results in the presence of other large datasets, more sophisticated mechanisms should be devised in terms of flexible parameter optimization.

Furthermore, we plan to design a more sophisticated GA hierarchy including containment relationships, and devise an algorithm that recognizes the complex GAs composed of multiple sub-GAs. Several existing studies \cite{chen2014,duong2005,fernando2016} use activity hierarchy for activity recognition. However, they usually construct a hierarchy from raw data level to event level or event level to complex activity (i.e.group activity in our paper) level. It is not possible to directly apply their approaches to our scheme because they do not have a hierarchy handling step for the same level and are inflexible to change the number of levels. We plan to incorporate activity hierarchy into ontology and recognize GAs in the hierarchy using the relationship semantics in the ontology. 

The proposed scheme has a limitation in recognizing multiple concurrent complex activities in a smart space. For example, a group of users performs \textit{Chatting} and another group of users does \textit{Playing a game} in the same lounge. Since we assume that users are anonymous due to privacy and accessibility issues, the data association problem \cite{hsu2008} occurs as well. Although recent studies \cite{Peng2018,Alemdar2017} have addressed multi-complex activity recognition, most of them include user information or use wearable sensors. Therefore, a study on multi-complex activity using only pervasive sensors is a natural extension to the proposed scheme and we plan to seek a solution to this by exploiting a multi-group checking process.

\section{Conclusion}
In this paper, we propose an efficient group activity recognition scheme based on causality-aware pattern mining in a pervasive sensor space to support high accuracy with missing or false data tolerance. 
To detect causally related event pairs and compute their temporal relations, we design a group activity ontology and a heuristics-based preprocessing method. For pattern learning, the proposed scheme sorts lists of preprocessed temporal-causal relations and extracts corresponding sequential patterns using the pattern-tree algorithm. For a comparative evaluation, we compare the recognition accuracy and runtime of the proposed scheme with three existing approaches, association rule mining, CRP-based ITBN, and HMM. The experiment results show that the proposed scheme outperforms the existing ones with the given dataset collected from our testbed while its run-time execution overhead does not heavily increase like CRP-based ITBN and HMM as the number of episodes gets large. In addition, to verify the proposed scheme is robust in real environments where data is missing or false values are generated, we conduct a noise analysis by adding noise data to training. The noise analysis result shows performance degradation ratio of the proposed scheme is smaller than that of the CRP-based ITBN scheme. We also prove, with the CASAS group activity dataset, the scalability of the proposed scheme in terms of applicability to various types of environments.

\bibliographystyle{ACM-Reference-Format}
\bibliography{bibliography}

\end{document}